\documentclass{article}

\usepackage{verbatim}
\usepackage{microtype}
\usepackage{graphicx}
\usepackage{amsmath, amssymb}
\usepackage{algorithm}
\usepackage{subfig}
\usepackage{booktabs} 
\usepackage[accepted]{icml2021} 
\usepackage{siunitx}
\usepackage[table,xcdraw]{xcolor}

\usepackage{listings}
\usepackage{hyperref}

\begin{document}

\twocolumn[
\icmltitle{Supervised Fine-Tuning or In-Context Learning? \\ Evaluating LLMs for Clinical NER}

\begin{icmlauthorlist}
\icmlauthor{Andrei Baroian}{} \\
LIACS, Leiden University
\end{icmlauthorlist}


\icmlcorrespondingauthor{Cieua Vvvvv}{c.vvvvv@googol.com}
\icmlcorrespondingauthor{Eee Pppp}{ep@eden.co.uk}
\vskip 0.3in
]

\begin{abstract}
We study clinical Named Entity Recognition (NER) on the CADEC corpus and compare three families of approaches: (i) BERT-style encoders (BERT Base, BioClinicalBERT, RoBERTa-large), (ii) GPT-4o used with few-shot in-context learning (ICL) under simple vs.\ complex prompts, and (iii) GPT-4o with supervised fine-tuning (SFT). All models are evaluated on standard NER metrics over CADEC’s five entity types (ADR, Drug, Disease, Symptom, Finding). RoBERTa-large and BioClinicalBERT offer limited improvements over BERT Base, showing the limit of these family of models. Among LLM settings, simple ICL outperforms a longer, instruction-heavy prompt, and SFT achieves the strongest overall performance (F1 $\approx$ 87.1\%), albeit with higher cost. We find that the LLM achieve higher accuracy on simplified tasks, restricting classification to 2 labels. 
\end{abstract}

\section{Introduction}

Introduction of electronic health records (EHRs) and the Internet has unlocked the access to a wealth of clinical information that could transform how data in medicine is handled. The sources of such data are clinical notes, patient forums, electronic health records (EHRs), and social media platforms. These data sources hold a wealth of information that, if effectively extracted and analyzed, can significantly enhance our understanding of patient experiences, treatment outcomes, and the safety profiles of medications. However, this unstructured format presents challenges in terms of accessibility and interpretability for automated systems.\cite{bethany2021}

Text mining, a key area within natural language processing (NLP), has emerged as a powerful tool to process and derive meaningful insights from unstructured textual data \cite{nadeau2007survey}. In clinical contexts, text mining applications span diverse tasks, such as identifying adverse drug reactions (ADRs), extracting symptoms and diseases from medical records, and mapping clinical concepts to standardized terminologies. Named Entity Recognition (NER), one of the foundational tasks in text mining, is instrumental in identifying and categorizing specific entities such as drugs, symptoms, or diseases within text. The integration of NER into clinical workflows has demonstrated potential in improving pharmacovigilance, enhancing clinical decision-making, and supporting medical research \cite{biomedical_NER}.

Despite these advancements, clinical text mining faces numerous challenges. Clinical narratives are often written in inconsistent formats, with extensive use of abbreviations, jargon, and ambiguous expressions \cite{bethany2021}. Additionally, data created as part of medical care is subject to rigorous scrutiny in terms of patient privacy \cite{gdpr}. This discourages researchers from publishing source data and decreases the availability of high-quality medical data. Patient forums  can theoretically provide a proxy for such data without the added threshold of privacy concerns.

This paper seeks to improve upon the state of the art of clinical Named Entity Recognition by leveraging the NER-annotated data from the Csiro Drug Adverse Effects Corpus (CADEC) \cite{CADEC2015} and robust transformer-based models.

\subsection{Research Questions}

The goal of the research is to explore new methods that have the potential to increase performance on Clinical NER task.

\textbf{Research Questions 1:
}
\begin{itemize}
    \item How does a domain-specific transformer model (BioClinicalBERT) compare to general-purpose transformer models (BERT Base) when applied to the CADEC corpus for Clinical NER?

\end{itemize}

\textbf{Research Questions 2}:

\begin{itemize}
    \item How do few-shot In-Context Learning and supervised fine-tuning of GPT-4 compare to BERT models in identifying clinical entities on the CADEC corpus?

\end{itemize}


\section{Background}

\subsection{Named Entity Recognition (NER) History}

The NER task was traditionally solved using supervised techniques including Hidden Markov Models (HMM), Support Vector Machines (SVM), Conditional Random Fields (CRF), with CONLL-2003 as one of the key benchmarks, and with complex linguistic \& multilingualism as critical challenges \cite{nadeau2007survey}. In recent years, researchers have adapted Deep
Learning techniques to effectively tackle the NER tasks, starting with simple Convolutional Neural Network (NN) and Recursive NN and advancing to Neural Language Models and Transform Models \cite{NERdl}. In their NER survey paper, \citet{NERsota2021} mention  BiLSTM-CRF (a hybrid model) as the state-of-the-art, but they fail to take Transformers into account. \citet{NERdl} acknowledge both methods and reveal that Transform methods (BERT) outperform BiLSTM-CRF on NER benchmarks. Both surveys mention the challenges of NER: scarcity of annotated datasets and informal \& noisy text.

\subsection{NER in Medical Context}

\citet{BioMedical_NER_perera2020named} summarise the developments of BioMedical NER, explaining the methods mentioned above and concluding that BioBERT is the state-of-the-art (SOTA) for NER in BioMedical Context.
A year later, \citet{biomedical_NER} examine and compare the performance of multiple NER methods on Bio Medical Datasets. Five methods (CRF, GRAM-CNN, Bi-LSTM-CRF, MTM-CW \& BioBERT) were compared over six datasets and BioBERT proved the best model overall (always better than BiLSTM-CR) with GRAM-CNN outperforming it on one datasets . A caveat that most of the datasets are inclined towards biology entities like \textit{Genes}, \textit{Protein}, \textit{Chemicals}, but there are datasets which takes \textit{Disease} as entities which makes this survey paper relevant to the current report. It matters as performance "vary considerably across different datasets"\cite{biomedical_NER}.
\citet{BioBERT_extension} acknowledge BioBERT as SOTA but also its "not always satisfactory" performance, and work to improve it by treating the problem as Machine Reading Comprehension (MRC) one instead of sequence labeling problem. They claim that this method obtain SOTA results, but looking at its performance, the improvement is rather marginal and the paper does not prove statistical significance differences.

The articles mentioned above target BioMedical NER which is closely related to Clinical NER. Other articles tackled Clinical NER, but they were not taken into account as they are relatively old (2018-2020) and in different languages than English (Spanish, Portuguese, Chinese) - thus not serving to the goal of exploring new methods to improve upon SOTA in Clinical NER.

\citet{Clinical_NER<3_2021} present a survey on both Clinical NER and RE (relationship extraction) and as the surveys presented above, they mention previous methods used and conclude that BioBERT consistently outperforms other methods for Clinical NER and RE tasks.

\subsection{NER and LLMs}

The above mentioned papers are came before the LLM era (starting 2022). Since then, researchers tried using LLM as to extract information, including NER from clinical text. 

\citet{LLMsurvey} present the NER and ER performances of different LLMs. Three paradigms were taken into account: In Context Learning (ICL), Data Augmentation (DA) and Supervised Fine Tuning (SFT). 
The following observations refer to the findings on 5 NER datasets:
\begin{itemize}
    \item EnTDA (with T5 as the base model and using the DA paradigm) yielded the best overall results and stability.
    \item BERT models are still competitive, performing similarly to or better than most LLMs, but are slightly outperformed by EnTDA.
    \item Supervised Fine-Tuning exhibits much better performance compared to In-Context Learning.
    \item Fine-tuned LLMs perform best on some datasets but have a higher variance in overall performance.
\end{itemize}


Concretely for Clinical NER, \citet{hu2024improvinglargelanguagemodels} showed that LLMs' (GPT 4) performance can be improved (from 0.804 F1 score to 0.861) only by prompt engineering in the ICL paradigm, but still underperforms compared to BioClinicalBERT. Four levels of prompts were used in the study: Baseline prompts, Annotation guideline-based prompts, Error analysis-based instructions and annotated samples via few-shot learning.

\section{Data Description}
CSIRO Adverse Drug Event Corpus (CADEC) \cite{CADEC2015} is a comprehensive collection of information from posts sourced from an on-line forum. The AskAPatient forum collects and allows for the exchange of information from patient reviews of medications.
Karimi et al. obtained 1250 posts, all in English, from AskAPatient and divided them into two categories based on the active ingredient of the reviewed medicine (Diclofenac and Lipitor). From the posts, they extracted the free text sections (removing sections like the rating or the demographic information). Subsequently, the researchers annotated the data in two steps: entity identification and terminology association (normalisation).

\begin{figure}[h!]
    \centering
    \includegraphics[width=\linewidth]{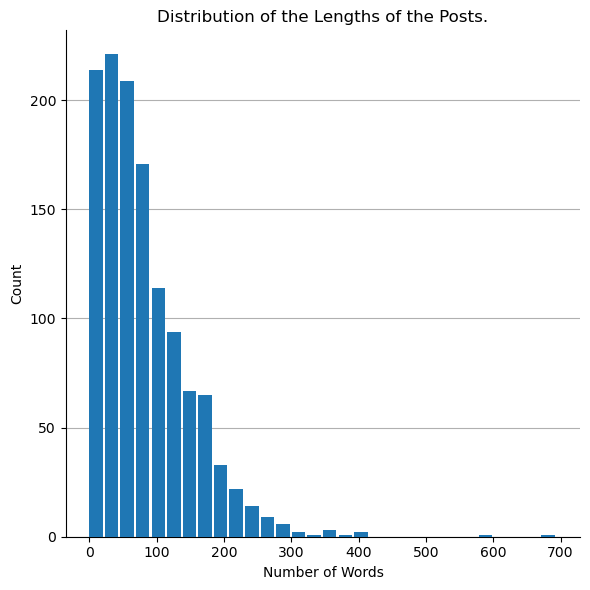}
    \caption{Distribution of the length (expressed as the number of words) of the posts that were included in CADEC.}
    \label{fig:post_text_length_distr}
\end{figure}
\subsection{Entity identification}
In this step, human annotators identified entities of interest in the texts. In order to achieve more reliable annotations, the researchers devised guidelines. An annotation could be discontinuous within a sentence, but it had to be contained within one sentence.

Annotation types distribution:
ADR: 6318
Drug: 1800
Finding: 435
Disease: 283
Symptom: 275

Types of annotated entities:
\begin{enumerate}
    \item \textbf{Drug}: Mention of a drug name (but not a drug class or a medical device). - 1800 instances
    \item \textbf{ADR}: Mention of an adverse drug effect along with the necessary context (e.g., 'felt blank'). - 6318 instances
    \item \textbf{Disease}: Mention of the disease that was the reason for taking the medicine in question. - 283 instances
    \item \textbf{Symptom}: A symptom that was the reason for taking the medicine in question. - 275 instances
    \item \textbf{Finding}: Any adverse side effect, disease or symptom that was not directly experienced by the reporting patient, or any other clinical concept that could fall into any of these categories but the annotator is not clear as to which one it belongs. - 435 Instances
\end{enumerate}

\begin{figure}[ht]
    \centering
    \includegraphics[width=.8\linewidth]{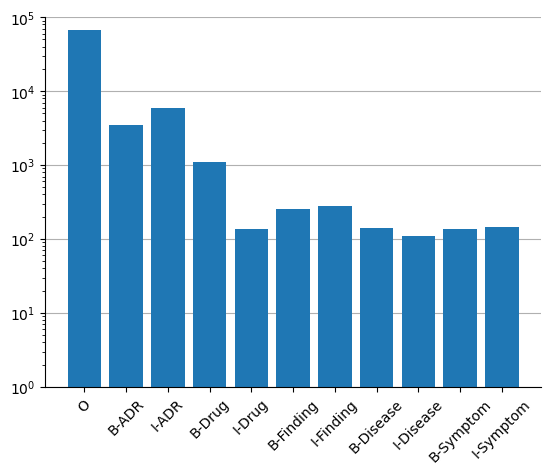}
    \caption{Distribution of the IOB entity labels in CADEC.}
    \label{fig:iob_distribution}
\end{figure}

\subsection{Terminology association (normalisation)}
In order to add informational value to the annotated entities, the researchers performed a terminology association step that normalises entities to their proper clinical meaning.
The process was performed using three terminologies: SNOMED CT, MedDRA and AHT.
All entities except Drug were mapped by a clinical terminologies to SNOMED CT.
Drug was mapped in a one-to-one manner to AHT. Finally MedDRA was used to map the SNOMED CT entities to a lowest level term (a most specific term that expresses an individual's condition).

\subsection{Final corpus form}
After the two annotation steps, the dataset was presented, representing each post using for files: (1) Raw source text. (2) Annotated entities, their types (e.g., \textit{Drug} or \textit{Symptom} and location in the source text. (3) SNOMED CT mapped entities and their location in the source text. (4) MedDRA mapped entities and their location in the source text.
The locations in the source text were expressed as the number of characters from the start and end of each part of said entity.

\subsection{Challenges of the dataset}
While the dataset of a relatively high quality, it does not come without its shortcomintgs. The source data used for the generation of the corpus was user-generated. User-generated data tends to be ambiguous, inconsistent, or even wrong. Additionally, such texts can obtain colloquialisms and less rigorous use of a language. Moreover, the size of the corpus is relatively small; with only 1250 posts, the dataset can be used for fine-tuning a pre-trained model but the results may vary. In addition to the general size limitation, the variety of the included data is quite low: the researchers focused on only two active substances (Diclofenac and Lipitor) which would make generalisation to more drug types and families difficult. Finally, human annotators were used to identify entities which, depending among other reasons on the quality of the guidelines, can yield to varying quality of the data.

\begin{figure}[h!]
    \centering
    \includegraphics[width=\linewidth]{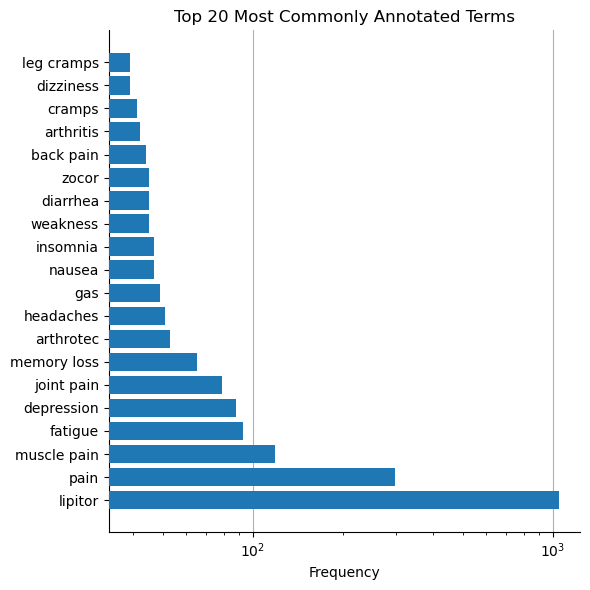}
    \caption{Top 20 most common terms in CADEC.}
    \label{fig:common_terms}
\end{figure}

\section{Methods}

To answer the research questions seven experiments were conducted. 
\citet{LLMsurvey} showed that for general NER tasks, Supervised Fine-Tuned LLMs provide excellent F1 scores, similar to SFT BERT models, while In Context Learning LLM exhibited inferior performance.  
Based on \citet{LLMsurvey}'s survey, the following models have been experimented on, to solve the Clinical NER task:

\begin{enumerate}
    \item BERT models
    \begin{enumerate}
        \item BERT Base - Baseline
        \item BioClinicalBERT
        \item RoBERTa-large
    \end{enumerate}
    \item In Context, few-shot learning GPT4o
    \begin{enumerate}
        \item Complex Prompt
        \item Simple Prompt
        \item Simple Prompt, 3 Labels 
    \end{enumerate}
    \item Supervised Fine-Tuning GPT4o (Complex Prompt)
\end{enumerate}

\textit{BERT Base} was chosen as a baseline, \textit{BioClinicalBERT} was chosen to see the importance of domain-specific knowledge and transfer learning, and \textit{RoBERTa-large} was chosen as the best BERT method to compare with the LLMs. Three experiments were made to find the best results for In Context, few-shot learning: First, a complex prompt including laborious details about the datasets and task with five full examples; Second, a simple prompt with few, essential details and small examples; Third, a simple prompt and testing only 3 labels (O, B-ADR, I-ADR) instead of all 11, to increase the simplicity of the task. ADR was chosen as it contains the highest number of instances. The reason for fine-tuning a closed sources LLM (GPT4o) rather than an open-source model is described in later sections. Expectations based on prior work are that RoBERTa-large and SFT GPT4o will perform best with similar performance.

\subsubsection*{Data Processing and Preparation}
Data Processing and Preparation are the same for all models. First, the text files and their corresponding annotations are processed using regular expressions to extract raw text and labels. Annotations are parsed to identify entity boundaries and types, with each annotation containing Named Entity Recognition (NER) labels and the annotated text.

\subsubsection*{Evaluation}

Evaluation is the same for all models. Model performance was evaluated using standard NER metrics including precision, recall, F1-score, and Evaluation Runtime (in seconds).  

\subsection{BERT models}

All three models \textit{BERT Base}, \textit{BioClinicalBERT} and \textit{RoBERTa-large} were fine-tuned for Clinical NER on the CADEC dataset using the same method, described below:

The Hugging Face Transformers library is used for implementation. The NER task was structured to identify the 11 entity labels (ADR, Drug, Finding, Disease, and Symptom entities), each with its corresponding \textit{B-} or \textit{I-} tag, plus an Outside ('O') tag.

\subsubsection*{Token and Label Generation}
The text processing uses regular expressions to identify words and punctuation. Proper token-label alignment is maintained with the help of a recursive boundary checking mechanism - it determines the appropriate tag for each token based on its position within annotated spans. The tokenization process uses BERT's byte-level (BPE) tokenizer.

\subsubsection*{Dataset and Training}
The implementation utilizes the Hugging Face \texttt{Dataset} and \texttt{DatasetDict} classes for efficient data management. The data is split into training (64\%), validation (16\%), and test (20\%).
 All models were trained on a learning rate of $2 \times 10^{-5}$, 5 epochs and 0.01 weight decay.

\subsubsection{BERT Base}

\citet{devlin2019bertpretrainingdeepbidirectional} introduced in 2019 BERT (Bidirectional Encoder Representations from Transformers), a transformer-based machine learning model pre-trained on a large corpus of English text. It has the ability to capture contextual understanding and proved it's value by achieving good performances on multiple benchmarks, including those for NER tasks. As an open-source model, it got many upgrades from researchers, from injecting domanin-specific knowledge to increasing efficiency.

\subsubsection{BioClinicalBERT}

Standford researchers \citet{alsentzer2019publiclyavailableclinicalbert}  introduced also in 2019 BioClinicalBERT, a domain-specific model developed to increase performance on NLP tasks in the clinical domain. BioClinicalBERT builds upon BioBERT, a variant of BERT pre-trained on biomedical texts, and further fine-tuned using clinical notes from a dataset containing more than 2 million electronic health records. 
The model was trained to address the linguistic and structural challenges of clinical texts, and  the experiments showed that BioClinicalBERT outperformed models like BERT in scenarios requiring domain-specific understanding. The question now is if CADEC dataset require that kind of understanding.

\subsubsection{RoBERTa Large}

RoBERTa (Robustly Optimized BERT Approach) \citet{RoBERTa} was created by FaceBookAI and released in 2019 (as well). RoBERTa addresses several limitations in BERT's original training process. Importnat modifications include training the model for longer periods of time with larger batche sizes, and dynamically masking input sequences during training. Even more, RoBERTa utilizes more training data, from diverse datasets. Evaluations on multiple benchmarks such as show better performance compared to the BERT model.

\subsection{Large Language Model NER}

\subsubsection*{Model Inference via GPT-4}
An OpenAI API client (GPT-4) is used to perform NER on the CADEC dataset. A user prompt is used to give the  input text and a system prompt assures adherence to the specified labeling and output rules. The prompts were inspired by \citet{hu2024improvinglargelanguagemodels} methods (See Appendix \ref{app:complex}). An early error of the model was the output of many 'O's (counted 913 for one example) after the prediction of the real text. This was solved by prompting the exact length of the sequence and asked it to adhere to it, but it was not efficient every time because of the inherited tokenization disadvantage of the LLMs of counting badly. 

The GPT-4 responses are parsed to extract predicted labels, which are further checked for IOB consistency. Lastly, the predicted labels are compared to true labels and evaluation metrics computed. 

\textbf{Budget} - because the OpenAI API was used and it requires credits, a budget of \$20 was set. 

\subsubsection{In Context, few-shot learning LLM}

To test the effect of prompt engineering on performance and to try to improving performance, three ICL LLM methods were used: Complex Prompt, Simple Prompt and Simple Prompt 3 Labels. The complex vs simple prompt should show if the LLM would perform better if there are fewer instructions, thus easier to follow and adhere to. Simple prompt, 3 labels aims at further simplifying the task by giving only 3 tags (O, B-ADR, I-ADR). If this method gets 100\% accuracy on this criteria, it would translate to an weighted accuracy of 0,693. If that would be the case, then 5 different models can be trained for each entity and ensembled into one output, potentially reaching better performance.

\subsubsection{Supervised Fine-Tuned LLM}

There are two options to fine-tune a large language model: open source and closed source, each with advantages and disadvantages. The Open source models offer flexibility -  more control over the hyperparameters - but they require high computational resources and are difficult to implement. On the contrast, closed source models are easy to utilize (upload supervised data) and do not require any resources, but they lack flexibility and charge money per token use. Giving the constraints, the closed-source models were preferred and ChatGPT4o was chosen.  

First, training dataset for SFT was created. The format must match those of a conversation between an user and an AI assistant, including a system prompt. Using the prompts from ICL LLM section and CADEC data, 100 conversations were created as training set. Although the training set for SFT BERT contained more than 1,000 examples, the recommendations found on OpenAI website \cite{openai_finetuning} and budget constraints limited the size of training dataset to 100. Then, a model was created using OpenAI developer's UI tool, although it was possible to do it in Python code. Once the model was fine-tuned, it was evaluated using the same approach as ICL LLMs.


\section{Results}

Table \ref{tab:ner_results} shows the Precision, Recall, F1 Score and Evaluation Runtime (in seconds) of each method.

\begin{table*}[h!]
\centering
\begin{tabular}{lcccc}
\hline
\textbf{Model} & \textbf{Precision} & \textbf{Recall} & \textbf{F1 Score} & \textbf{Eval Runtime (s)} \\
\hline
BERT Base & 0.5533 & 0.6669 & 0.6048 & 2.096 \\
BioClinicalBERT & 0.5632 & 0.6857 & 0.6184 & 41.74 \\
RoBERTa-large & 0.5980 & 0.6991 & 0.6446 & 143.251 \\
ICL 3 Labels GPT4o & \textcolor{gray}{0.869} & \textcolor{gray}{0.8871} & \textcolor{gray}{0.8748} & \textcolor{gray}{360.549} \\
Weigthed ICL 3 Labels & 0.6067 & 0.603 & 0.615 & 520.273\\
ICL Simple GPT4o & 0.8353 & 0.8261 & 0.8306 & 409.042 \\
ICL Complex GPT4o & 0.7290 & 0.8538 & 0.7865 & 805.273 \\
\textbf{SFT Complex GPT4o} &\textbf{ 0.8534 }& \textbf{0.8894} & \textbf{0.8710} & \textbf{818.818} \\
\hline
\end{tabular}
\caption{NER performance comparison across different model approaches. }
\label{tab:ner_results}
\end{table*}

For ICL 3 Labels GPT4o, the metrics are strictly on ADR and not on all 5 entities, thus, weighted averaging is used to compute overall metrics, taking TP,FP and F1 score equal to 0 for other entities and averaging over the entire dataset distribution (ADR instances = 6318, All instances = 9111). 
The same weighted average is performed on Evaluation Runtime. These transformations offer only approximations and might differ from the true values.

\small
\[
F1_{\text{weighted}} = \frac{6318 \times 0.8748 + 0 + 0 + 0 + 0}{9111} 
\approx \frac{5530}{9111} 
\approx 0.6067.
\]

\small
\[
Precision
_{\text{weighted}} \approx \frac{6318 \times 0.8692}{9111} \approx 0.603, \quad
\]

\small
\[
Recall_{\text{weighted}} \approx \frac{6318 \times 0.8871}{9111} \approx 0.615.
\]

\small
\[
T_{\text{total}} \approx \frac{360.5494}{\frac{6318}{9111}} \approx \frac{360.5494}{0.693} \approx 520.1 \, \text{seconds}.
\]

\subsection{Result Analysis}

\subsubsection{BERT models}

As expected \textit{BERT Base} was the fastest and obtained a decent F1 Score, representing a good baseline. Surprisingly \textit{BioClinicalBERT} did not have a significant effect on performance (1.36\% improvement) while taking 20 times more to compute, suggesting that extra domain-specific knowledge is not required. An even bigger surprise was the performance of \textit{RoBERTa-large}, only 3\% better than Bert Base with 70 times more compute. The surprise come from the fact that in \citet{LLMsurvey}, it had excellent results, similar to SFT LLMs. A reason for poor performance might be the lack of hyperparameter tuning, which was not used to make the comparison fair. Future research should apply hyperparameter optimization to all three BERT methods in order to get the most out of this models and be comparable to LLMs.

\subsubsection{LLMs}

Comparing Simple and Complex prompts for ICL GPT4o, simple prompts yieled better results (3\% increase in F1 score) and proved more efficient (half the runtime). Looking at the \ref{app:complex}, it can be observed the lengthy examples might be contra-productive and confuse the model rather than enhance it. Even so, the SFT Complex GPT4o exhibited the best performance, and showed 6\% improvement over ICL version, using almost the same eval runtime. A caveat is that the evaluation runtime does not take into account fine-tuning time or costs. The improvement was expected, although in \citet{LLMsurvey}'s survey, they found greater discrepancies between ICL and SFT. Future research should try SFT on the Simpler Prompt. 

Lastly, the most simplified LLM method, ICL 3 Labels, exhibited the highest F1 Score but it takes into account only ADR entity when computing the metric. There is a possibility that if 5 models are trained, their ensemble would yield the same performance as ICL 3 Labels and prove the best method for Clinical NER. But until then, the weighted average is much lower than other LLM methods.

\citet{SoTA_CADEC} claimed in 2022 to have found State-of-the-art performance on CADEC with F1 score of 86.69\%, but only on ADR entity, and F1 Score of 78.76\% when including all entities. The performances from our experiments are better than previously claimed SoTA. This statement should be taken with the caveat that those results were found two years ago.

\section{Discussion}

\subsection{Verifying the results}

In order to verify if the methods were implemented correctly and if the results hold, another literature review was done to check performances of the same or similar models on the same dataset - CADEC Corpus.

\citet{yuan2022biobartpretrainingevaluationbiomedical} got an F1 score of 68.37 for Bart Base, and 68.39 for BioBERT Base.
An in-depth analysis of different models on CADEC corpus was performed by cite \citet{CADEC_compare}. They found the BERT model yield a Strict F1 Score of 66.67\%, RoBERTa (not large) 65.83\% and BioClinicalBERT 66.23\%. This shows a high discrepancy in performance and raises concerns whether the results are valid. Future research to repeat the experiments. Even so, the conclusion regarding domain-specific knowledge holds strong.

Only a few academic articles tested large language models on CADEC dataset, but only on open models like GPT-2 or Llam-7B. Thus, there is no guarantee of the validity of the results.

\subsection{Future Directions}

Even if the current results are promising, there is a probability combining the insights found in this report will improve the performance significantly. Future research should use Supervised Fine-Tuning on Simple Prompt, and on the whole dataset. A direction that could be added to the above suggestions, is to ensemble/combine 5 models, once for each entity. Another direction is to try implement Data Augumentation to enhance performance.

To improve credibility of results and method implementation, future research should obtain results for established NER datasets using the same models. If the performance would be similar to the findings of other academic articles, it would confirm the results on CADEC dataset. 

\subsection{Limitations}

One on the biggest limitation is that only one type, closed-sourced LLM was used for both ICL and SFT, which limits the flexibility, adds extra costs and limits the generalizability to other LLMs. Future research should aim fine-tuning multiple open source models.  Another limitation is the lack of hyperparameter optimization for all models which might inhibit the performance.

\section{Conclusion}

This study aimed at finding best-performing model for the Clinical NER task on CADEC corpus. Prior research shows good performance for SFT BERT models and SFT LLMs. Seven experiments were conducted: three on BERT models, three on In Context Learning GPT4o and one on Supervised Fine-Tuned GPT4o. The results claim that SFT GPT4o produced the best F1 Score of 87.1 \% but at the same time incurred the highest costs. The research suggests that performance can be enhanced further by fine-tuning on the entire training dataset and using a simpler prompt. The study's limitations are in the number of models tested, the number of datasets the models are tested on, and the lack of hyperparameter tuning which might enhance performance.

\bibliographystyle{icml2021} 
\bibliography{main}

\section{Appendix}

\appendix
\section{PROMPTS}

\subsection{Complex Prompt}
\label{app:complex}
\subsubsection{System Prompt}
\begin{lstlisting}
### Task
Your task is to identify and label named entities in medical text using IOB (Inside-Outside-Beginning) notation. The entities to be identified are: ADR (Adverse Drug Reactions), Drug, Finding, Disease, and Symptom.

### Entity Markup Guide
- B- prefix indicates the Beginning of an entity
- I- prefix indicates Inside (continuation) of an entity
- O indicates Outside (not part of any entity)
- Labels: B-ADR, I-ADR, B-Drug, I-Drug, B-Finding, I-Finding, B-Disease, I-Disease, B-Symptom, I-Symptom, O

### Annotation Guidelines
1. ADR (Adverse Drug Reactions):
   - Mark symptoms/conditions that occur as reactions to medications
   - Include complete phrases describing the reaction
   - Example: "severe headache after taking aspirin" -> B-ADR I-ADR I-ADR

2. Drug:
   - Mark medication names and their variants
   - Include brand names and generic names
   - Example: "Lipitor" -> B-Drug

3. Finding/Disease/Symptom:
   - Finding: Medical observations or test results
   - Disease: Diagnosed medical conditions
   - Symptom: Patient-reported health issues not classified as ADRs

### Error Prevention Guidelines
Common mistakes to avoid:
1. Don't mark general health terms as entities
2. Distinguish between ADRs and pre-existing conditions
3. Don't split single entity mentions
4. Maintain consistency in multi-token entities
6. Review your final tags to ensure you never place an I-ADR label immediately after an O label, unless that token continues a multi-word ADR entity that was previously labeled
7. Provide only one tag per word, in order. If there are more tags than words, the extra ones will be discarded. So do not generate them

### Examples

Example 1:
Input:"" fatigue , muscle c ##ram ##mps , b ##loat ##ing , and intense chest pains . Also depression ( on second try ) . After one week , realized that I could not handle this ! I ' d rather live off rice cakes and water . I Wen ##t on a bike ride and felt like I had the flu . Wen ##t home , got on the ' net and searched for side effects and was amazed at what I found . Wen ##t off lip ##itor , but my do ##c convinced me to try again . After 3 days , same symptoms plus depression and mood swings . Stop ##ped again , started watching my diet more closely and have dropped my ch ##ole ##stro ##l by 30 p ##ts w / in a couple of months .""
Output:"" 'B-ADR', 'O', 'B-ADR', 'I-ADR', 'I-ADR', 'I-ADR', 'O', 'B-ADR', 'I-ADR', 'I-ADR', 'O', 'O', 'B-ADR', 'I-ADR', 'I-ADR', 'O', 'O', 'B-ADR', 'O', 'O', 'O', 'O', 'O', 'O', 'O', 'O', 'O', 'O', 'O', 'O', 'O', 'O', 'O', 'O', 'O', 'O', 'O', 'O', 'O', 'O', 'O', 'O', 'O', 'O', 'O', 'O', 'O', 'O', 'O', 'O', 'O', 'O', 'O', 'O', 'O', 'O', 'O', 'O', 'O', 'O', 'B-Disease', 'O', 'O', 'O', 'O', 'O', 'O', 'O', 'O', 'O', 'O', 'O', 'O', 'O', 'O', 'O', 'O', 'O', 'O', 'O', 'O', 'O', 'O', 'O', 'O', 'O', 'O', 'B-Drug', 'I-Drug', 'O', 'O', 'O', 'O', 'O', 'O', 'O', 'O', 'O', 'O', 'O', 'O', 'O', 'O', 'O', 'O', 'O', 'O', 'B-ADR', 'O', 'B-ADR', 'I-ADR', 'O', 'O', 'O', 'O', 'O', 'O', 'O', 'O', 'O', 'O', 'O', 'O', 'O', 'O', 'O', 'O', 'O', 'O', 'O', 'O', 'O', 'O', 'O', 'O', 'O', 'O', 'O', 'O', 'O', 'O', 'O'""

Example 2:
Input: ""After 1 year on Li ##pit ##or , I experienced severe pain in both hips & legs , would wake up with so much pain at night that I would get out of bed for relief . Even after 3 months of non ##use , I still have stiff ##ness , weakness & pain in both legs / hips , especially right side & now limp . I have started physical therapy , massage the ##raphy & pain medicine .""
Output:"" 'O', 'O', 'O', 'O', 'B-Drug', 'I-Drug', 'I-Drug', 'O', 'O', 'O', 'B-ADR', 'I-ADR', 'I-ADR', 'O', 'I-ADR', 'O', 'I-ADR', 'O', 'O', 'O', 'O', 'O', 'B-ADR', 'I-ADR', 'I-ADR', 'O', 'O', 'O', 'O', 'O', 'O', 'O', 'O', 'O', 'O', 'O', 'O', 'O', 'O', 'O', 'O', 'O', 'O', 'O', 'O', 'O', 'O', 'O', 'B-ADR', 'I-ADR', 'O', 'O', 'O', 'O', 'I-ADR', 'I-ADR', 'I-ADR', 'O', 'I-ADR', 'O', 'O', 'O', 'O', 'O', 'O', 'B-ADR', 'O', 'O', 'O', 'O', 'O', 'O', 'O', 'O', 'O', 'O', 'O', 'O', 'O'""

Example 3:
Input:"" My boyfriend is 49 and has had high ch ##ole ##ster ##ol for about 10 years . his doctor put him on Li ##pit ##or . he took it twice and could hardly stay awake . Also noticed pains in his legs and weakness in his knees after the first dose . He has stopped and will not resume this treatment . There must be a better alternative . Don ' t do it ! There are too many websites like these with pages and pages of similar complaints .""
Output:"" 'O', 'O', 'O', 'O', 'O', 'O', 'O', 'O', 'O', 'O', 'O', 'O', 'O', 'O', 'O', 'O', 'O', 'O', 'O', 'O', 'O', 'O', 'B-Drug', 'I-Drug', 'I-Drug', 'O', 'O', 'O', 'O', 'O', 'O', 'O', 'O', 'O', 'O', 'O', 'O', 'O', 'B-ADR', 'I-ADR', 'I-ADR', 'I-ADR', 'O', 'B-ADR', 'I-ADR', 'I-ADR', 'I-ADR', 'O', 'O', 'O', 'O', 'O', 'O', 'O', 'O', 'O', 'O', 'O', 'O', 'O', 'O', 'O', 'O', 'O', 'O', 'O', 'O', 'O', 'O', 'O', 'O', 'O', 'O', 'O', 'O', 'O', 'O', 'O', 'O', 'O', 'O', 'O', 'O', 'O', 'O', 'O', 'O', 'O', 'O', 'O'""

Example 4:
Input:  State ##d with joint and pain and muscle weakness , depression , fatigue and c ##ram ##ps . At each annual visit to the Dr I would tell him my symptoms and mention I thought it was the Li ##pit ##or , He wa ##ou ##ld always say no way and I need to continue the s ##tat ##in drug . My symptoms gradual ##y got worse over the next 5 - 6 years , currently I have muscle twitch ##ing in both arms , loss of muscle mass in arms , hands and face with s ##lu ##rred speech . Too ##k myself off the lip ##itor a year ago . Joint pain is com ##ple ##tly gone . Was told my Dr I may have AL ##S . Saw an AL ##S Specialist Feb , 2002 , she said she could not diagnosis AL ##S at this time as you have to have 3 ex ##ter ##mit ##ies effect ##ed and I have 1 1 / 2 . She said she has seen over 100 patients with symptoms the same as mine that had been on s ##tat ##in drugs . I can ' t begin to tell you what a scary year this has been . But am holding out hope I will improve in time . Have read that some people it ta ##ks 1 - 2 years to improve at all . I trusted my Doctor ! ! ! ! ! ! ! ! Now I trust that what I was fell ##ing 6 years ago was not all in my head . Anyone with si ##mu ##lar symptoms or come ##nts please e - mail .
Output: 'O', 'O', 'O', 'B-ADR', 'O', 'I-ADR', 'O', 'B-ADR', 'I-ADR', 'O', 'B-ADR', 'O', 'B-ADR', 'O', 'B-ADR', 'I-ADR', 'I-ADR', 'O', 'O', 'O', 'O', 'O', 'O', 'O', 'O', 'O', 'O', 'O', 'O', 'O', 'O', 'O', 'O', 'O', 'O', 'O', 'O', 'O', 'B-Drug', 'I-Drug', 'I-Drug', 'O', 'O', 'O', 'O', 'O', 'O', 'O', 'O', 'O', 'O', 'O', 'O', 'O', 'O', 'O', 'O', 'O', 'O', 'O', 'O', 'O', 'O', 'O', 'O', 'O', 'O', 'O', 'O', 'O', 'O', 'O', 'O', 'O', 'O', 'O', 'O', 'O', 'B-ADR', 'I-ADR', 'I-ADR', 'I-ADR', 'I-ADR', 'I-ADR', 'O', 'B-ADR', 'I-ADR', 'I-ADR', 'I-ADR', 'I-ADR', 'O', 'O', 'I-ADR', 'O', 'I-ADR', 'O', 'B-ADR', 'I-ADR', 'I-ADR', 'I-ADR', 'O', 'O', 'O', 'O', 'O', 'O', 'B-Drug', 'I-Drug', 'O', 'O', 'O', 'O', 'B-ADR', 'I-ADR', 'O', 'O', 'O', 'O', 'O', 'O', 'O', 'O', 'O', 'O', 'O', 'O', 'O', 'B-Finding', 'I-Finding', 'O', 'O', 'O', 'B-Finding', 'I-Finding', 'O', 'O', 'O', 'O', 'O', 'O', 'O', 'O', 'O', 'O', 'O', 'B-Finding', 'I-Finding', 'O', 'O', 'O', 'O', 'O', 'O', 'O', 'O', 'O', 'O', 'O', 'O', 'O', 'O', 'O', 'O', 'O', 'O', 'O', 'O', 'O', 'O', 'O', 'O', 'O', 'O', 'O', 'O', 'O', 'O', 'O', 'O', 'O', 'O', 'O', 'O', 'O', 'O', 'O', 'O', 'O', 'O', 'O', 'O', 'O', 'O', 'O', 'O', 'O', 'O', 'O', 'O', 'O', 'O', 'O', 'O', 'O', 'O', 'O', 'O', 'O', 'O', 'O', 'O', 'O', 'O', 'O', 'O', 'O', 'O', 'O', 'O', 'O', 'O', 'O', 'O', 'O', 'O', 'O', 'O', 'O', 'O', 'O', 'O', 'O', 'O', 'O', 'O', 'O', 'O', 'O', 'O', 'O', 'O', 'O', 'O', 'O', 'O', 'O', 'O', 'O', 'O', 'O', 'O', 'O', 'O', 'O', 'O', 'O', 'O', 'O', 'O', 'O', 'O', 'O', 'O', 'O', 'O', 'O', 'O', 'O', 'O', 'O', 'O', 'O', 'O', 'O', 'O', 'O', 'O', 'O', 'O', 'O', 'O', 'O'

Example 5:
Input: I ' ve really had no side affects what so ever . I have blood work done re ##gua ##lar ##ly to check my ch ##ole ##ster ##ol levels . I ' ve had high ch ##ole ##ster ##ol since I was 19 , now I ' m 38 and have been on this medication since 1997 . My ch ##ole ##ster ##ol is still high but not as high . In Canada the rating should be below 5 . 20 , with being on this medication its still about 6 . 24 were in 1997 my level was 8 . 45 . Its come down quite a bit , but I do exercise and watch what I eat as well .
Output: 'O', 'O', 'O', 'O', 'O', 'O', 'O', 'O', 'O', 'O', 'O', 'O', 'O', 'O', 'O', 'O', 'O', 'O', 'O', 'O', 'O', 'O', 'O', 'O', 'O', 'O', 'O', 'O', 'O', 'O', 'O', 'O', 'O', 'O', 'B-Finding', 'I-Finding', 'I-Finding', 'I-Finding', 'I-Finding', 'O', 'O', 'O', 'O', 'O', 'O', 'O', 'O', 'O', 'O', 'O', 'O', 'O', 'O', 'O', 'O', 'O', 'O', 'O', 'O', 'B-Finding', 'I-Finding', 'I-Finding', 'I-Finding', 'I-Finding', 'O', 'I-Finding', 'O', 'O', 'O', 'O', 'O', 'O', 'O', 'O', 'O', 'O', 'O', 'O', 'O', 'O', 'O', 'O', 'O', 'O', 'O', 'O', 'O', 'O', 'O', 'O', 'O', 'O', 'O', 'O', 'O', 'O', 'O', 'O', 'O', 'O', 'O', 'O', 'O', 'O', 'O', 'O', 'O', 'O', 'O', 'O', 'O', 'O', 'O', 'O', 'O', 'O', 'O', 'O', 'O', 'O', 'O', 'O'

\end{lstlisting}

\subsubsection{User Prompt}
\begin{lstlisting}
Your task is to identify and label named entities in medical text using IOB (Inside-Outside-Beginning) notation. There are {num_words} tokens in the text. Provide exactly {num_words} IOB labels. Do not add extra tokens beyond {num_words}. Input: {text}

Output: 
\end{lstlisting}

\subsection{Simple Prompt}

\subsubsection{System Prompt}

\begin{lstlisting}
    
You are an expert annotator. For each token in the user text, label it as one of:
- B-Drug if it starts a drug mention
- I-Drug if it continues a drug mention
- B-ADR if it starts an adverse drug reaction mention
- I-ADR if it continues an adverse drug reaction mention
- B-Disease if it starts a disease mention
- I-Disease if it continues a disease mention
- B-Symptom if it starts a symptom mention
- I-Symptom if it continues a symptom mention
- B-Finding if it starts a finding mention
- I-Finding if it continues a finding mention
- O otherwise

Rules:
1. If a mention has multiple words, use B- tag on the first word, I- tag on the rest.
2. Only use the specified labels above. No other labels.
3. Do not produce more labels than tokens. Output exactly one label per token.
4. If you are uncertain, label as O.

Examples:
Text: "mild headache from Lipitor"
Output: B-Symptom I-Symptom O B-Drug

Text: "diabetes caused severe joint pain"
Output: B-Disease O B-ADR I-ADR I-ADR

Text: "doctor found high blood pressure"
Output: O B-Finding I-Finding I-Finding
\end{lstlisting}

\subsubsection{User Prompt}

\begin{lstlisting}
The text has {len(words)} tokens.
Output exactly {len(words)} labels (one per token) using any of these labels:
B-Drug, I-Drug, B-ADR, I-ADR, B-Disease, I-Disease, B-Symptom, I-Symptom, B-Finding, I-Finding, O
Text: {text}

Output:    
\end{lstlisting}
  
\end{document}